\documentclass[10pt, a4paper]{article}
\usepackage{lrec}
\usepackage{multibib}
\newcites{languageresource}{Language Resources}
\usepackage{graphicx}
\usepackage{tabularx}
\usepackage{soul}
\usepackage{pifont}
\usepackage{epstopdf}
\usepackage{flushend}
\usepackage[T1]{fontenc}
\usepackage[utf8]{inputenc}
\usepackage{graphicx}
\usepackage{hyperref}
\usepackage{xstring}
\usepackage{eurosym}
\usepackage{mathpazo}
\usepackage{textcomp}
\usepackage{enumerate}
\usepackage{multirow}

\title{Tools and resources for Romanian text-to-speech and speech-to-text applications}

\name{Tiberiu Boros, Stefan Daniel Dumitrescu and Vasile Pais}

\address{Research Institute for Artificial Intelligence ``Mihai Draganescu'', Romanian Academy \\
         Calea 13 Septembrie, nr. 13 \\
         \{tibi, sdumitrescu, pais\}@racai.ro\\}

\abstract{
In this paper we introduce a set of resources and tools aimed at providing support for natural language processing, text-to-speech synthesis and speech recognition for Romanian. While the tools are general purpose and can be used for any language (we successfully trained our system for more than 50 languages and participated in the Universal Dependencies Shared Task), the resources are only relevant for Romanian language processing.\\
\newline  }

\begin{document}

\maketitleabstract

\section{Introduction}

Natural language processing (NLP), text-to-speech synthesis (TTS) and automatic speech recognition (ASR) are key components of modern applications especially those that rely on human-computer-interaction via voice input/output. As smart phones and gadgets are gaining ground on their competitors (laptops and desktops), they are the most likely candidates to serve as front-ends in the Internet-of-Thing (IoT) landscape. However, these devices rely mostly on low-powered chips that need to run live/responsive user interfaces. Our work is focused in providing support for NLP, TTS and ASR applications in low-resourced environments by providing a set of tools that is designed to easily scale, depending on the available application and computational resources. Given the success of the widely spread and well-known lightweight ASR system PocketSphinx \cite{huggins2006pocketsphinx} we only address NLP and TTS with our tools. We do however, introduce a newly created text-to-speech synthesis corpus that is intended to consolidate currently available speech resources for the Romanian language and a newly created speech recognition corpus which is composed of a freely available sub-corpus and license-restricted one. Though we are able to provide transcription and phoneme-level alignments for the non-free section of the ASR corpus, obtaining the recorded speech is the subject of a different process that involves a third party (the RADOR\footnote{http://www.rador.ro/} press agency). As we are currently working on a neural-based speech recognition and keyword spotting tool for Romanian, providing pre-trained models on the entire speech corpus will not be a problem and will mitigate the license restrictions. Also, we have to mention that the audio data can be indirectly obtained by systematically using the Oral Corpus Query Platform (OCQP) (later described in section \ref{ocqp}) from the COROLA project\footnote{http://corola.racai.ro/\#interogare} based on our aligned data.

The paper is structured as follows: (a) the first part introduces two ready-to-use frameworks that are freely available for download with no license restrictions; (b) the second part describes a freely available speech corpus for Romanian, focusing on corpora-composition and annotations; (c) the third part discusses our road-map for future developments and enhancements.

\section{Tools description}

\subsection{Natural Language Processing}

Most natural language processing (NLP) tasks require a certain level of text preprocessing aimed at segmenting the input into standard processing units (often into sentences and words but, depending on the application, also syllables, phonemes etc.) and at enriching these units with additional features designed to reduce the effect of data sparsity (lemmas, part-of-speech tags, morphological attributes etc.). Because this is a ground-zero requirement, the literature is abundant with methods and techniques for low-level text processing, but \textbf{multilingual text-processing is still a challenging task}. This has been proven by the well-known shared task on Universal Dependencies (UD) parsing \cite{udst:overview}. One very important conclusion is that while some algorithms have an overall better performance than others - and we draw the attention to Stanford's \cite{dozat2017parsing} graph-based parser, there is \textbf{no ``one size fits all'' algorithm} that is language and corpora-size independent. 

While accuracy carries a great weight in NLP applications, there are two other factors that impact the design of such systems: computational cost and memory footprint. With this in mind we introduced support for three very different machine learning algorithms applied on the same set of text-processing tasks: decision trees, linear models and neural networks (bidirectional long-short-term-memory (LSTM) networks). We motivate our choice based on the computational/memory requirements of these algorithms:
\begin{itemize}
\item \textbf{Decision trees} (DTs) require virtually \textbf{no feature-engineering}, provide a \textbf{relatively small model footprint}, with a \textbf{logarithmic computational complexity} ($O\left ( \log n \right )$), where $n$ is the number of unique features and \textbf{low mathematical load};
\item \textbf{Linear models} require \textbf{extensive feature-engineering}, yield models with \textbf{large footprints}, with \textbf{linear computational complexity} ($O\left ( n \right )$) and a \textbf{moderate mathematical load} (commonly multiplications and additions);
\item \textbf{Neural networks} are able to learn patterns and automatically generate required non-linearities between the input features, yield \textbf{small footprint models} (even with compact feature embeddings), but generate a \textbf{high computational load}, mainly because of the large number of operations and the use of \textbf{complex mathematical functions} (multiplications, additions, $\tanh$ or $\sigma$ activation functions).
\end{itemize}

\begin{figure}[!h]
\begin{center}
\includegraphics[width=0.47\textwidth]{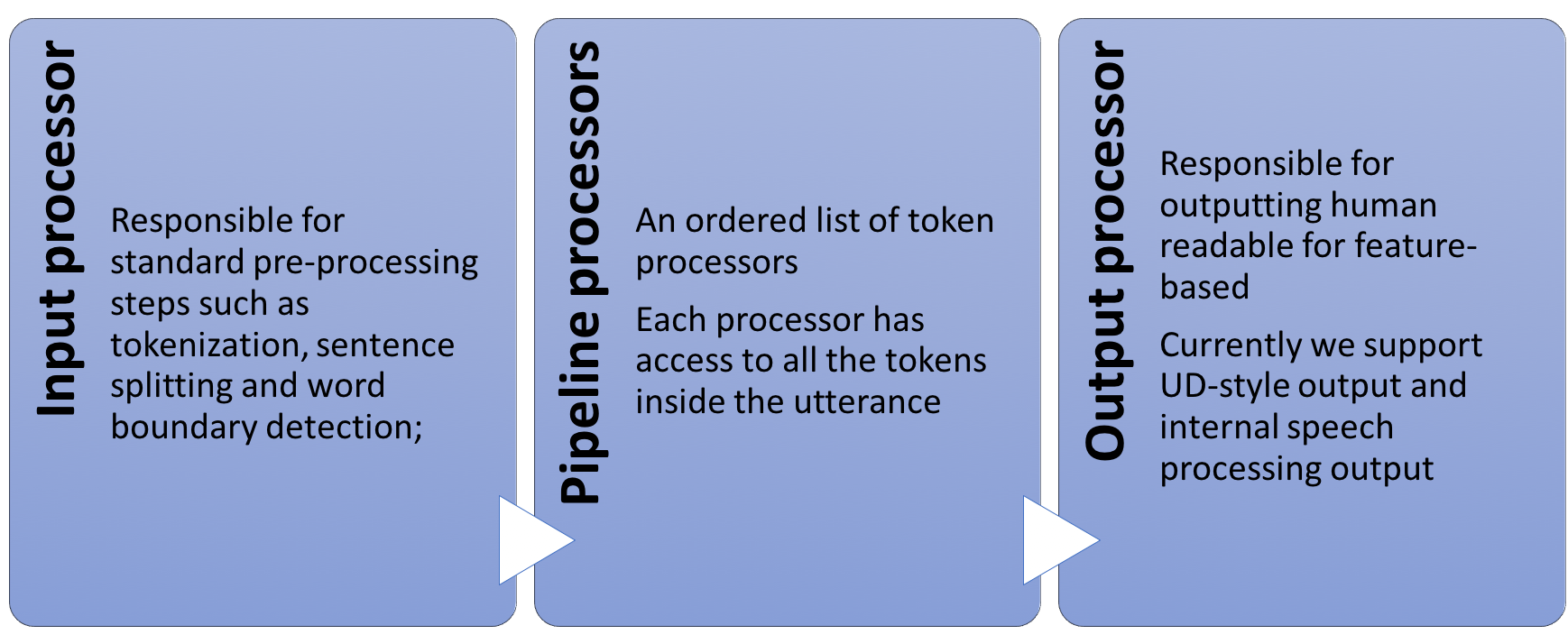} 
\caption{Overview of the MLPLA architecture}
\label{figure:architecture}
\end{center}
\end{figure}

The MLPLA architecture (Figure \ref{figure:architecture}) (initially introduced in \cite{zafiu2015modular}) is composed of three main layers: (a) input; (b) processing pipeline and (c) output. For overcoming language-dependent and approach-based limitations, our system is built using a scalable plug-and-play methodology. The processing units are built so that they implement one of three different interfaces depending on the module under which they operate. The three interfaces are (a) the data input processor - an implementation of this interface must be able to receive the input text as a character sequence and perform any necessary preprocessing in order to obtain a tokenized text; (b) the base processor interface – an implementation receives a sequence of tokens, each token containing standard attributes (part-of-speech, lemma, phonetic transcription, syllables, accent, chunk and dependencies) but also allowing the insertion of non-standard attributes through a key-value table - each processor is responsible for building its own feature set using all available data, performing the NLP task it was designed for and filling in either a value for a standard attribute or adding a custom attribute; (c) the feature-based output – an implementation must take a sequence of tokens and convert it into a feature-based output, depending on the application in which MLPLA is used. The order in which the base processors are chained is controlled externally from a configuration file (see Figure \ref{fig:config} for details).
\begin{figure}
\centering
\begin{tabularx}{0.47\textwidth}{l}
\hline
\textbf{{[}Input{]}} \\
mlpla.language.preprocessing.BasicTokenizer \\
\textbf{{[}Pipeline{]}} \\
mlpla.language.baseprocessors.BasicTagger \\
mlpla.language.baseprocessors.BasicLemmatizer \\
mlpla.language.baseprocessors.BasicChunker \\
mlpla.language.baseprocessors.BasicParser \\
mlpla.language.baseprocessors.BasicSyllabifier \\
mlpla.language.baseprocessors.BasicLTS \\
mlpla.language.baseprocessors.BasicStress \\
\textbf{{[}Output{]}} \\
mlpla.language.formats.TabFeatureOutput \\ \hline
\end{tabularx}
\caption{Excerpt from the MLPLA configuration file}
\label{fig:config}
\end{figure}

Our recent work has been centered on extending the existing system and addressing multilingual text processing. Given that we are able to easily interchange between models/modules and classifiers, we focused our efforts into assessing what is the best trade-off between speed/accuracy and model size because versatility is an important feature of our framework. 

\textbf{NOTE:} The performance of each model processing steps are currently not the focus of this paper. However, they can be looked up in \newcite{zafiu2015modular} for Romanian specific data and in \newcite{dumitrescu2017universal} for the complete list of languages supported by UD. If this article is accepted as a long paper we plan to include more results using a stacked bidirectional LSTM model that we've worked on recently.

We note that the system supports tokenization, lemmatization, chunking, part-of-speech tagging, parsing, syllabification, stress prediction on words and letter-to-sound (for text-to-speech purposes). Each of these processing tasks uses one or more of the algorithms shortly described below; for example, tagging can be done either with a linear model or with LSTMs. The configuration file allows easy prototyping of solutions using our platform.

\subsection{Text-to-speech synthesis}
The speech synthesis tool is called SSLA which stands for Speech Synthesis for Lightweight Applications. We implemented statistical parametric speech synthesis because it offers constant quality and a small footprint in contrast to the concatenative (unit-selection) approach that might sound more natural at times (if enough data is available, otherwise worse than parametric synthesis) and a significantly larger memory requirement. We use decision trees to independently model frame-by-frame speech parameters for the spectral envelope, phone-state duration and voice pitch. 

For the effective speech synthesis process it can switch between the classical Mel-Log Spectral Approximation (MLSA) filter \cite{imai1983mel} and Speech Transformation and Representation using Adaptive Interpolation of weiGHTed spectrum (STRAIGHT) \cite{kawahara1999restructuring}. The reason for this is that while STRAIGHT results are superior in quality, it is way more computationally expensive that the MLSA counterpart and, for some applications this can be a really important bottleneck. 

The input models are fully compatible with the HMM Speech Synthesis (HTS) Toolkit \cite{zen2007hmm}, which in fact we use for training and compiling models. Currently we only support multinomial training and we treat features as ``bag-of-words''. There are no constraints regarding the feature format, except that features should not contain spaces or the special character `/' which is the feature delimiter.

The standard feature-set used by our speech synthesis back-end is:
\begin{itemize}
\item \textbf{Phonetic context}: (a) current phoneme accompanied by two preceding phonemes and two succeeding phonemes; (b) articulatory information for all phones inside the feature window;
\item \textbf{Syllable information}: (a) the identity of frequent syllables\footnote{In our experiments we set the threshold to 5 for frequent syllables} is used as a feature along side with (b) information which is present regardless of the syllables frequency such as: lexical accent, relative syllable position inside the word and sentence, the total number of syllables in the sentence (which is actually a feature used for the global variance), distance from the previous and distance to the next punctuation mark;
\item {Global context information}: the type of the sentence (declarative, interrogative or exclamation), the identity of the previous and the next punctuation marks and the total number of words inside the utterance;
\item {Local morphosyntactic information}: previous, current and next part-of-speech together with the relative position of the syllable and word inside syntactic chunks as defined in \newcite{ion2007word} 
\end{itemize}

\subsection {Oral Corpus Query Platform}
\label{ocqp}

The Oral Corpus Query Platform is an on-line tool designed to help linguists in their study of spoken language. It enables one to query our oral corpus using combinations of wordforms, lemmas and part-of-speech tags. It is currently part of the COROLA project \cite{tufis2015corola} and it is hosted in the RACAI cloud\footnote{http://korap.racai.ro/corola\_sound\_search/index.php}, but, if desired we are willing to provide access to our code-base and help deploy the platform on-site. 

The data currently available on this platform contains the Romanian oral corpus which is described in the next section, as well as additional speech corpora from the Institute of Computer Science of the Romanian Academy (IIT). 

In order to fully support indexing and searching through the corpus we used the flat start monophones procedure of HTK \cite{young2002htk} in order to obtain phoneme-level alignments between the transcriptions and speech data. Because HTK only uses words and their phonetic transcriptions we realigned the raw text data with the phoneme-level information using dynamic programming. Also, the raw text data was tokenized, lemmatized and tagged using an external tool called TTL \cite{ion2007ttl}. The reason for not using our own tool-chain was that COROLA required consistent annotations over the entire corpora and the text-component was already processed using this standard tool.

\section {Speech corpus}
As previously mentioned our speech data is composed of a section aimed at text-to-speech synthesis (composed of high-quality recordings) and another section which is intended to provide support for speech recognition applications.
\subsection {The text-to-speech synthesis corpus}
\subsubsection{Corpora composition}
One of the prerequisites in developing a TTS corpus states that the corpus must provide a good coverage over the target language and domain. In other words this means that (a) the corpus must be phonetically balanced in terms of target speech units (i.e. phonemes, diphones etc.) and (b) a single unit must appear in multiple prosodic contexts in order to enable the TTS system to learn the prosodic patterns that relate to the language, the target domain and the speaking style of the speaker himself. 
Taking into consideration the above mentioned conditions we decided to construct a Romanian speech corpus composed of two sections: 
\begin{enumerate}[(a)]
\item The first section (section A) is based on Wikipedia (for Romanian) and contains a number of sentences that were chosen using a greedy algorithm (that will be presented later in this paper) in order to ensure the completeness of the phonetic domain of the Romanian language. The sentences are treated as individual prompts (no larger context is provided), thus the speaker must record each individual sentence “out of the blue” and he is forced to limit his narrative interpretation to the utterance itself.
\item The second section of the corpus (section B) is composed of the Romanian adaptation after Allen Carr’s book ``Easy way to stop smoking''. The book contains a lot of motivational and persuasive passages which are carefully crafted by the author to convince smokers quit their habit. Additional to the prompts themselves, we also made use of an existing audiobook. Originally, this audiobook was recorded by a male actor and has approximately two and a half hours of high quality studio recordings at 48KHz. This lead the actor to make use of highly prosodic rhetoric speech with the purpose of (a) reshaping the cognitive state of (b) and relying embedded messages to the listener. Gaining access to the low-level prosodic parameters (F0, phone duration and pauses) that make up such a speech is an asset to research in the field of natural TTS systems. The matching prompts (from the audiobook) were made available to our speakers (one male and one female) in order to act as a baseline and a guide in their voice shaping process.
\begin{table*}[]
\centering
\caption{Individual speaker statistics extracted from the phoneme-level aligned speech corpus}
\label{table:tts}
\begin{tabular}{c|rr|rr|rr|rr}
\hline\multicolumn{1}{c|}{ } & \multicolumn{2}{c|}{\textbf{Speaker 1 (female)}}                       & \multicolumn{2}{c|}{\textbf{Speaker 2 (female)}} & \multicolumn{2}{c|}{\textbf{Speaker 3 (male)}} & \multicolumn{2}{c}{\textbf{Speaker 4 (male)}} \\
\textbf{Phoneme}       & \textbf{Occ} & \textbf{Tot. length} & \textbf{Occ}       & \textbf{Tot. length}       & \textbf{Occ}      & \textbf{Tot. length}      & \textbf{Occ}      & \textbf{Tot. length}      \\\hline
\textbf{@}             & 3301         & 225150               & 3963               & 278330                     & 4529              & 277930                    & 2540              & 177530                    \\
\textbf{a}             & 10942        & 951670               & 14932              & 1376420                    & 17279             & 1384770                   & 5004              & 409820                    \\
\textbf{a@}            & 1754         & 91750                & 2097               & 123630                     & 2376              & 130300                    & 1094              & 67540                     \\
\textbf{b}             & 992          & 73270                & 1490               & 124170                     & 1682              & 133220                    & 365               & 26670                     \\
\textbf{ch}            & 1539         & 164230               & 1965               & 227920                     & 2206              & 262139                    & 839               & 72760                     \\
\textbf{d}             & 3624         & 213640               & 4897               & 318940                     & 5652              & 362660                    & 1946              & 127410                    \\
\textbf{dz}            & 332          & 32940                & 634                & 70390                      & 731               & 78530                     & 125               & 9740                      \\
\textbf{e}             & 11701        & 770900               & 14678              & 976310                     & 16991             & 1053870                   & 5390              & 362330                    \\
\textbf{e@}            & 1042         & 54360                & 1364               & 70680                      & 1598              & 81770                     & 564               & 32550                     \\
\textbf{f}             & 1482         & 138980               & 1734               & 185380                     & 2012              & 216239                    & 925               & 85220                     \\
\textbf{g}             & 904          & 60610                & 1193               & 98320                      & 1376              & 102870                    & 436               & 31750                     \\
\textbf{h}             & 280          & 29320                & 393                & 46449                      & 488               & 61060                     & 43                & 3210                      \\
\textbf{i}             & 7596         & 484880               & 9737               & 685550                     & 11240             & 718580                    & 3436              & 235340                    \\
\textbf{ij}            & 1655         & 76460                & 2028               & 111459                     & 2388              & 140870                    & 693               & 44200                     \\
\textbf{j}             & 2030         & 133060               & 2943               & 204670                     & 3366              & 260040                    & 1042              & 72040                     \\
\textbf{k}             & 4418         & 368310               & 5473               & 436100                     & 6373              & 530070                    & 2128              & 152020                    \\
\textbf{l}             & 5139         & 277920               & 7111               & 400320                     & 8172              & 476540                    & 1850              & 101680                    \\
\textbf{m}             & 2985         & 228550               & 3800               & 313770                     & 4382              & 370190                    & 2057              & 145020                    \\
\textbf{n}             & 6886         & 378840               & 9191               & 482750                     & 10657             & 655190                    & 3286              & 187470                    \\
\textbf{o}             & 4471         & 354630               & 5794               & 450240                     & 6631              & 442470                    & 1804              & 146520                    \\
\textbf{o@}            & 413          & 27270                & 491                & 35310                      & 600               & 33560                     & 227               & 13200                     \\
\textbf{p}             & 3394         & 266380               & 4246               & 364239                     & 4894              & 424840                    & 1481              & 114340                    \\
\textbf{pau}           & 1985         & 526943               & 2186               & 3373752                    & 2490              & 3883761                   & 1412              & 1541130                   \\
\textbf{r}             & 7965         & 351600               & 10428              & 483170                     & 12079             & 569870                    & 3200              & 137040                    \\
\textbf{s}             & 4351         & 424560               & 5672               & 575170                     & 6525              & 739510                    & 2096              & 183390                    \\
\textbf{sh}            & 1145         & 122640               & 1543               & 180770                     & 1848              & 235920                    & 702               & 64010                     \\
\textbf{sp}            & 5216         & 243840               & 3739               & 736331                     & 5023              & 1029306                   & 1700              & 137580                    \\
\textbf{t}             & 7045         & 498910               & 8613               & 619090                     & 9965              & 729590                    & 3380              & 231820                    \\
\textbf{ts}            & 1365         & 136910               & 1640               & 171210                     & 1895              & 205560                    & 705               & 59370                     \\
\textbf{u}             & 5855         & 359180               & 7680               & 474569                     & 8876              & 464020                    & 2822              & 166780                    \\
\textbf{v}             & 1427         & 97560                & 1641               & 143560                     & 1906              & 139430                    & 711               & 47320                     \\
\textbf{w}             & 703          & 58710                & 846                & 74960                      & 1000              & 87060                     & 113               & 9610                      \\
\textbf{z}             & 922          & 85960                & 1107               & 103760                     & 1242              & 123720                    & 453               & 39990                     \\
\textbf{zh}            & 410          & 44910                & 475                & 56680                      & 551               & 69340                     & 89                & 8470                      \\\hline
\textbf{Total (h)} & \multicolumn{2}{r}{\textbf{2.32}}   & \multicolumn{2}{r}{\textbf{4}}                           & \multicolumn{2}{r}{\textbf{4.58}}                      & \multicolumn{2}{r}{\textbf{1.46}}                     \\
\textbf{Overall (h)}       & \multicolumn{8}{c}{\textbf{12.36}}                                                                           \\\hline                                                                        
\end{tabular}
\end{table*}

\end{enumerate}
The second section of our corpus (the book section) is not as well balanced as the first section. The corpus from which section A sentences were extracted was the full dump of the Romanian Wikipedia as of June 2012, because, belonging to the encyclopedic genre, it contains a wide range of domains and different word types. 

Because the Wikipedia dump contains a lot of errors and is far from a clearly readable text, we had to employ a number of heuristic rules to remove and/or correct sentences. Below we enumerate the processing steps applied:
\begin{enumerate}[(a)]
\item Sentence-split the corpus and tokenize it, keeping only the ones that were not longer than 20 words. Using our in-house developed sentence splitter (based on a Maximum Entropy engine), we obtained over 5 million such sentences. 
\item Remove all leading and trailing spaces or non-printable characters.
\item Remove all lines that contain any of the following characters: `\textonehalf', `\ding{108}', `\textthreequarters', `\ding{109}', brackets, slashes, quotes, etc. (several characters we manually input), as well as all the lines that contain abbreviations or tokens like : Sos., Cal., .ro., uk., www., etc. . All these rules were input manually because there is a large number of sentences that contain there tokens and are not suitable for recording. Some of the rules are regexes like a word having Latin a-z characters; others were simple conditions that a line should not have a certain substring.
\item Remove all lines that contain numbers.
\item Remove all lines that are all caps (usually titles)
\item Remove all lines with less than three words with the following exceptions: if the sentence length is one, then that word should be in the Romanian Lexicon, thus removing a significant number of foreign sentences existing in Wikipedia.
\item Remove all lines that do not have at least 90\% words in the lexicon (excepting proper nouns). This rule ensured that a lot of erroneous sentences were removed because they contained words in foreign languages (even though we used the Romanian Wikipedia dump, we still found a great number of sentences that are or at least contain words in other languages).
\item Remove all lines that do not have at least 90\% words with diacritics, skipping the majority of existing foreign sentences.
\item Correct the Romanian i-of-i (î) words to the correct form of i-of-a (â). For example, the old word form ``cîte'' (meaning ``how many'') was corrected to the new writing ``câte''. While deterministic, this process is not straightforward relying on a lexicon, backing off to a specific set of rules that involve word decomposition.
\end{enumerate}
Step by step, the number of sentences decreased to approx. 252000 (only ~19\% of sentences passed the cleaning and correction phase). Interestingly, most lines that were removed were because they had numbers (d) or did not contain the minimum percent of words in the lexicon (g). On this set of sentences we applied the triphones balancing algorithm described next.
To keep the number of triphones from each type as balanced as possible (a perfect balancing is not possible because there are triphones that are intrinsically rare) we have applied the following algorithm:
\begin{enumerate}
\item Compile an initial frequency of triphones from the whole corpus;
\item If a sentence contained a rare triphones (with a frequency below 100), keep it;
\item If a sentence contained only very frequent triphones (with frequencies over the H index of the initial distribution), discard it.
\item Default action: keep the sentence.
\item Finally, sort the sentences according to the least common triphones first: this will ensure a balanced corpus from the start, no matter how many sentences we record out of the entire corpus.
\end{enumerate}

\subsubsection {Recording details}
The corpus was recorded in studio conditions by two professional speakers (male and female). This speech corpus is freely available for download and use. It is composed of 6h:30m:23s (female speakers) and 6h:03m:46s (male speakers) and the archive contains the speech prompts (one file each), corresponding audio files, phonetic transcription lexicon and time-aligned phoneme sequences for each prompt-audio pair. Table \ref{table:tts} shows a quantitative evaluation of the speech corpus.

For a qualitative evaluation, we provide statistical parametric speech synthesis models that are compatible with our platform, both for the STRAIGHT and MLSA vocoders. In the near future we intend to extend our speech synthesis platform to support WORLD \cite{morise2016world} for real-time high-quality vocoding and we will include pre-trained models as well. Currently SSLA can be queried on-line\footnote{http://slp.racai.ro/index.php/ssla/} for speech synthesis using one male and one female voice.

\subsection{The speech recognition corpus}
As earlier stated, the speech recognition corpus is composed of two subsections: the non-free sections which contains recordings from the RADOR agency and a collection of audio-books provided by IIT and the free-section which was internally created based on volunteers who recorded utterances from a predefined set of data. The quality of the recordings varies within the entire speech corpus, from sampling rate to noise conditions. The lowest recording sample rate is 16Khz and the highest is 48Khz. In terms of recording conditions we have studio recordings, semi-studio recordings (high quality equipments but no hemianechoic room) and standard desktop/laptop/headset recording equipment in noisy environments.

The corpus is sentence-split and each sentence is time-aligned with the speech data at phoneme level. Also we keep internal an internal-track of the source and recording conditions for every sentence. However, in this paper we will only provide quantitative information regarding the corpora composition divided between the two sections: free and non-free. 

\begin{table*}[]
\centering
\caption{Phoneme distribution and duration for the two sections of the ASR corpus: free and non-free}
\label{table:asr}
\begin{tabular}{c|rrr|rrr}
\hline
\multirow{2}{*}{\textbf{Phoneme}} & \multicolumn{3}{c}{\textbf{Non-free}}                                                                            & \multicolumn{3}{c}{\textbf{Free}}                                                                                \\
                         & \multicolumn{1}{c}{\textbf{Occurences}} & \multicolumn{1}{c}{\textbf{Total duration}} & \multicolumn{1}{c}{\textbf{Mean dur.}} & \multicolumn{1}{c}{\textbf{Occurences}} & \multicolumn{1}{c}{\textbf{Total duration}} & \multicolumn{1}{c}{\textbf{Mean dur.}} \\\hline
\textbf{@}	&	52117	&	4108212	&	78,83	&	73516	&	5580501	&	75,91 \\
\textbf{a}	&	168665	&	14646891	&	86,84	&	248419	&	21268426	&	85,62 \\
\textbf{a@}	&	25805	&	2158373	&	83,64	&	36451	&	2896279	&	79,46 \\
\textbf{b}	&	13971	&	872960	&	62,48	&	20620	&	1302030	&	63,14 \\
\textbf{ch}	&	26430	&	2431859	&	92,01	&	38491	&	3522853	&	91,52 \\
\textbf{d}	&	58951	&	3436241	&	58,29	&	85621	&	5032471	&	58,78 \\
\textbf{dz}	&	4062	&	332270	&	81,80	&	5985	&	478950	&	80,03 \\
\textbf{e}	&	186060	&	13077767	&	70,29	&	271792	&	18861070	&	69,40 \\
\textbf{e@}	&	14495	&	638588	&	44,06	&	21093	&	934677	&	44,31 \\
\textbf{f}	&	17927	&	1548020	&	86,35	&	26928	&	2253210	&	83,68 \\
\textbf{g}	&	10674	&	657900	&	61,64	&	15900	&	991005	&	62,33 \\
\textbf{h}	&	1559	&	117870	&	75,61	&	2259	&	165015	&	73,05 \\
\textbf{i}	&	109493	&	6623230	&	60,49	&	163033	&	9928170	&	60,90 \\
\textbf{ij}	&	30917	&	1949659	&	63,06	&	44094	&	2680152	&	60,78 \\
\textbf{j}	&	32366	&	2138951	&	66,09	&	47568	&	3135555	&	65,92 \\
\textbf{k}	&	64171	&	4671150	&	72,79	&	92329	&	6786150	&	73,50 \\
\textbf{l}	&	75989	&	3479229	&	45,79	&	113536	&	5122048	&	45,11 \\
\textbf{m}	&	52091	&	3614139	&	69,38	&	74572	&	5106568	&	68,48 \\
\textbf{n}	&	109934	&	5741698	&	52,23	&	159700	&	8360862	&	52,35 \\
\textbf{o}	&	69089	&	5166650	&	74,78	&	102393	&	7529220	&	73,53 \\
\textbf{o@}	&	6956	&	370200	&	53,22	&	9781	&	481665	&	49,24 \\
\textbf{p}	&	53568	&	3810840	&	71,14	&	78439	&	5634345	&	71,83 \\
\textbf{pau}	&	21672	&	6198915	&	286,03	&	33459	&	10481797	&	313,27 \\
\textbf{r}	&	119344	&	5049519	&	42,31	&	176946	&	7288335	&	41,19 \\
\textbf{s}	&	70045	&	6348160	&	90,63	&	101028	&	8991750	&	89,00 \\ 
\textbf{sh}	&	20623	&	2118431	&	102,72	&	29284	&	2957326	&	100,99 \\
\textbf{sp}	&	50862	&	12363894	&	243,09	&	72142	&	15154882	&	210,07 \\
\textbf{t}	&	109401	&	7008668	&	64,06	&	160231	&	10399795	&	64,91 \\
\textbf{ts}	&	19940	&	1811200	&	90,83	&	29230	&	2653155	&	90,77 \\
\textbf{u}	&	88979	&	5529020	&	62,14	&	130866	&	7798155	&	59,59 \\
\textbf{v}	&	21825	&	1309280	&	59,99	&	31038	&	1862353	&	60,00 \\
\textbf{w}	&	11053	&	881050	&	79,71	&	16323	&	1225815	&	75,10 \\
\textbf{z}	&	15409	&	1203250	&	78,09	&	23503	&	1807170	&	76,89 \\
\textbf{zh}	&	3654	&	335310	&	91,77	&	5407	&	489360	&	90,50 \\
\hline
\textbf{Total (hours)}            & \multicolumn{3}{c}{\textbf{36.59}}                                                                                    & \multicolumn{3}{c}{\textbf{52.54}}  \\
\textbf{Overall (hours)}            & \multicolumn{6}{c}{\textbf{89.14}}                                                                           
\\\hline                                    
\end{tabular}
\end{table*}

The corpus construction is still an on-going work. Aside from the data described in Table \ref{table:asr} we will enhance the free section of the corpus with at least another 20 hours of speech data, which is currently being processed. 

As mentioned, the data varies in quality across the entire speech corpus. In order to test if this corpus is relevant at all for automatic speech recognition we constructed a \textbf{character-level } (not phoneme-level) speech recognition system which uses Mel-generalized cepstral coefficients extracted using a 5-ms sliding window, which are fed into a two layer bidirectional LSTM (400 cells in each direction -- total 800 cells per layer) on top of which we use a softmax layer, trained using Connectionist Temporal Classification (CTC) loss\cite{graves2013speech}. This system architecture combined with a RNN language model for word segmentation will be fully described in our future work. However, we must state that after 4 training epochs on the entire training dataset, we obtained a \textbf{character-level accuracy rate} of 89.52\%. To our knowledge, this is the only character-level speech recognition system for Romanian and the results, which are consistent with those reported for other languages, show that this corpus can indeed be used to train ASR systems.

Whereas we are unable to say anything about the fidelity of the transcriptions for the non-free section, our speech data is carefully crafted and the error count is surely low. Additionally, our transcriptions take into account recoding and speech artifacts (noise, laughter, caught etc.) as well as foreign words (which are transliterated) and regional accents (for which we account by introducing the academic form of the word and the transliterated version that follows the actual pronunciation).

\section {Future development plans}

There are three main directions we want to proceed to in the near future: (a) extension of the tool-set; (b) pre-training models and (c) creation of additional resources for Romanian.

\textbf{Extension of the tool-set}: For the NLP module we seek to introduce a graph-based dependency parser which uses a complex network architecture composed of stacked Bidirectional LSTMs for feature extraction and a multilayer perceptron for word-arc scoring, similar to the approaches proposed in \newcite{dozat2016deep} and \newcite{kiperwasser2016simple}. Also, based on the success of deep-learning applied to TTS \cite{oord2016wavenet} we plan to extend our speech synthesis back-end to include neural speech synthesis support;

\textbf{Pre-trained models}: Depending on the language and training corpora size and composition, all models require some fine-tunning, weather we are talking about model hyper-parameters for neural-networks or feature-combinations for linear models. As such, we plan to provide pre-trained NLP models for all languages included in the Universal Dependencies Treebank \cite{ud20data}

\textbf{Text-to-speech corpora}: During our subjective internal evaluation of the speech models we noticed that the TTS system had a poor quality (in terms of prosody) when used in dialogue-style conversations. Intuitively, this is because neither the Wikipedia section nor the Audiobook section did include short dialogue sentences in our speech corpus. However, this type of interaction is typical for assistive systems, thus our future development plans include the extension of the speech corpus and inclusion of short dialogue sentences. 

\textbf{Speech recognition corpora}: As already mentioned, we are still working on extending our speech recognition corpus with new data.   

\section {Conclusions}

We have presented two ready-to-use tools and a speech resource that enable to construction and deployment of NLP and TTS applications in low-resourced environments. Of course, every component is independent and can be used in a standalone scenario to provide functionality (NLP or TTS tool) or to be used as input in training other systems.

The speech corpus is intended for Romanian, but the tools can be trained for any language. In fact, our demo shows how we trained NLP support for more than 50 languages \footnote{http://slp.racai.ro/mlpla-new}. 

The tool set is available for download (code and binaries)\footnote{http://slp.racai.ro/} and was tested both on desktop/server environments as well as on mobile devices (Android 5 and 6).

Furthermore, on request, we are happy to provide more pre-trained TTS and NLP models that are not currently available on the website.

\section*{Acknowledgments}
The corpora construction work described in this paper was supported though the the Heimdallr project, which is funded by the Romanian Government through the Executive Agency for Higher Education, Research, Development and Innovation Funding (UEFISCDI), programme "Experimental demonstration project (PED) PED-2016", project ID: PN-III-P2-2.1-PED-2016-1974, contract number 229PED.

We also want to thank all contributors and volunteers who supported us through recordings performed on the Romanian Anonymous Speech Corpus (RASC)\footnote{http://rasc.racai.ro/} platform.

\section{Bibliographical References}
\label{main:ref}

\bibliographystyle{lrec}
\bibliography{main}

\end{document}